\documentclass[sn-mathphys]{sn-jnl}

\jyear{2021}%

\theoremstyle{thmstyleone}%
\theoremstyle{thmstyletwo}%

\theoremstyle{thmstylethree}%

\begin{document}

\title[Multimodal Query-guided Object Localization]{Multimodal Query-guided Object Localization}


\author[1]{\fnm{Aditay} \sur{Tripathi}}\email{aditayt@iisc.ac.in}

\author[1]{\fnm{Rajath R} \sur{Dani}}\email{rajathrdani@gmail.com}

\author[2]{\fnm{Anand} \sur{Mishra}}\email{mishra@iitj.ac.in}

\author*[1]{\fnm{Anirban} \sur{Chakraborty}}\email{anirban@iisc.ac.in}
\affil[1]{\orgdiv{CDS}, \orgname{Indian Institute of Science}, \orgaddress{ \city{Bengaluru}, \postcode{560012}, \state{Karnataka}, \country{India}}}

\affil[2]{\orgdiv{CSE}, \orgname{Indian Institute of Technology}, \orgaddress{\city{Jodhpur}, \postcode{342037}, \state{Rajasthan}, \country{India}}}





\abstract{\textcolor{black}{Consider a scenario in one-shot query-guided object localization where neither an image of the object nor the object category name is available as a query. In such a scenario, a hand-drawn sketch of the object could be a choice for a query. However, hand-drawn crude sketches alone, when used as queries, might be ambiguous for object localization, e.g., a sketch of a laptop could be confused for a sofa. On the other hand,  a linguistic definition of the category, e.g., ``\textit{a small portable computer small enough to use in your lap}" along with the sketch query, gives better visual and semantic cues for object localization.}
In this work, we present a multimodal query-guided object localization approach under the challenging open-set setting. In particular, we use queries from two modalities, namely, hand-drawn sketch and description of the object (also known as gloss), to perform object localization. Multimodal query-guided object localization is a challenging task, especially when a large domain gap exists between the queries and the natural images, as well as due to the challenge of combining the complementary and minimal information present across the queries. For example, hand-drawn crude sketches contain abstract shape information of an object, while the text descriptions often capture partial semantic information about a given object category. To address the aforementioned challenges, we present a novel cross-modal attention scheme that guides the region proposal network to generate object proposals relevant to the input queries and a novel orthogonal projection-based proposal scoring technique that scores each proposal with respect to the queries, thereby yielding the final localization results. Our method is useful towards effectively localizing multiple object instances present in the target image not only in the challenging open-set setting (when some object categories are not seen during training) but also in the traditional closed-set case, where samples from all the object categories are available during training. In our experiments, hand-drawn sketches from the `Quick, Draw!' dataset and gloss retrieved from `WordNet' are used as queries to evaluate the localization performance of our approach on widely-used object localization public benchmark MS-COCO. The proposed method clearly outperforms all related baseline approaches across both closed-set and open-set object localization tasks.}

\keywords{Sketch, Open-set object localization, Gloss, Cross-modal localization, Cross-modal Attention}



\maketitle

\section{Introduction}
\label{sec: intro}
We have seen breakthroughs in object detection literature in the last decade, and it is partly due to the advancements in deep learning~\citep{ren2016faster,liu2016ssd,redmon2018yolov3,lin2017feature}. However, most of these successful models are still limited to `closed-world' settings, where the object localization and classification tasks are limited to a predefined set of categories whose examples are used during the training phase. In this work, we study a more challenging task of open-set query-guided object localization with the following goal -- given an image of a natural scene and an object query, localize all the instances of the queried object in the image, even if no sample for this queried object is assumed available during the training phase. \textcolor{black}{In the literature, query-guided object localization has been attempted using either object category name~\citep{WangRHT14} or an image of the object as a query~\citep{hsieh2019one,sivic2003video}. However, it is possible to encounter scenarios where neither an image of the object nor the class label is available as a query. Such a scenario can arise (i) due to privacy reasons or (ii) when the object of interest is uncommon (not a natural object, e.g., parts of a machine). However, even in such a scenario, it is often easy to find a crude drawing or natural language description of the query object. We, therefore, want to explore the following task-\emph{Can a hand-drawn sketch or natural language description of any object be used for localizing all the instances of the corresponding object in a natural scene?} We investigate the answer to this research question in this work.} 
 
\begin{figure}[!t]
    \centering
    \includegraphics[width = \textwidth]{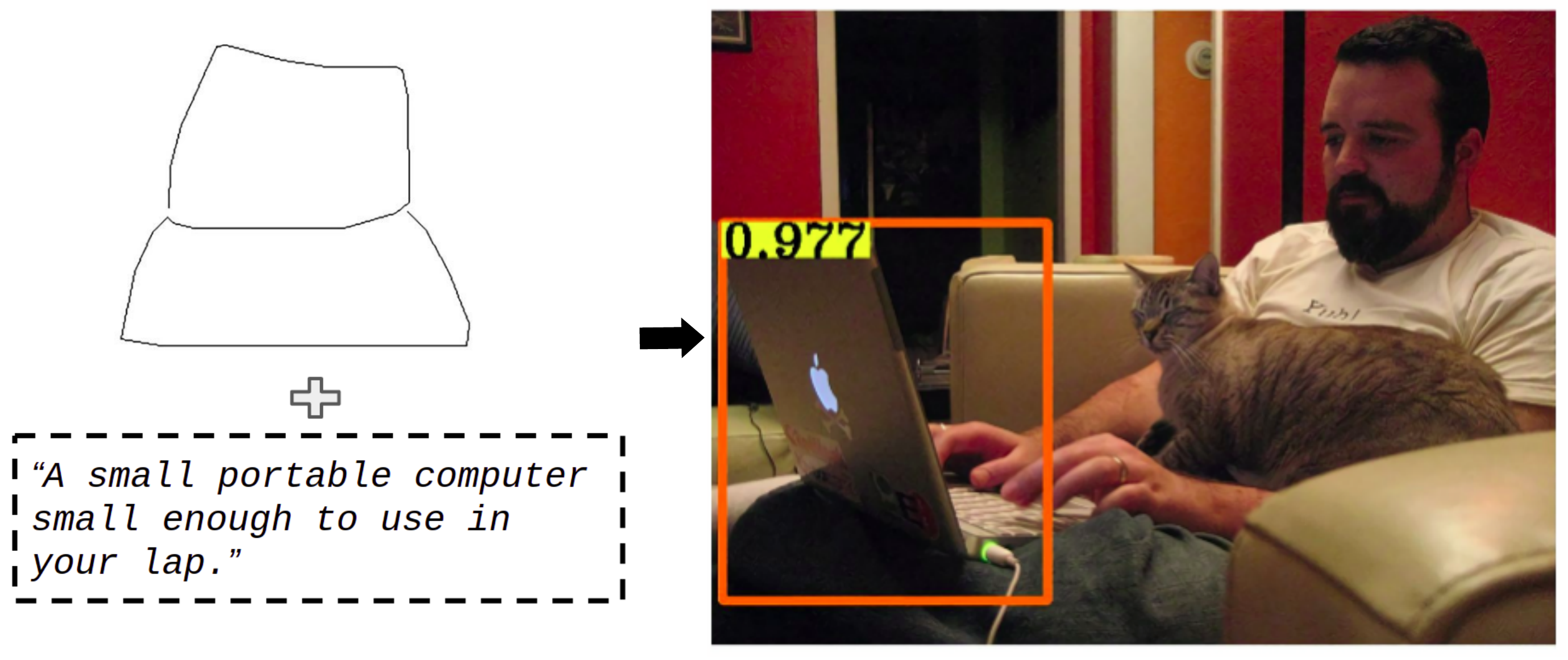}
    \caption{\label{fig:goal}  Given an image and a query, our aim is to localize the object in the image  (a laptop in this example). A hand-drawn sketch of a laptop alone, when used as a query, might be ambiguous for object localization as it could be confused for a sofa. On the other hand, descriptions obtained from different modalities such as a category label, e.g. ``laptop" or a linguistic definition of the category, e.g., ``a small portable computer small enough to use in your lap" along with the sketch query, give better visual and semantic cues for the object localization.}
\end{figure}

In our earlier work~\citep{tripathi2020sketch}, we introduced the novel idea of using hand-drawn sketches of objects as queries towards localizing objects in a natural scene. Sketches provide an abstract visual representation of the objects. Most free-hand sketches (e.g., sketches in Quick Draw) lack serious visual content, such as the appearance, color, and texture of the drawn objects. Often, these sketches only provide noisy outlines depicting the global shapes of an object and lack any finer structural details. For example, consider Figure~\ref{fig:goal}, where a hand-drawn sketch of a laptop might be ambiguous for object localization as it could be confused for a sofa. \textcolor{black}{These, understandably, lead to very limited success in the sketch-guided object localization task. On the contrary, by combining different modalities, such as visual (in the form of a hand-drawn sketch of the query object) and text (in the form of a natural language description of the query object), it is possible to leverage complementary and intricate details on an object's shape, appearance or texture and sometimes even the semantic relationship of the query with other objects in the scene.} Judiciously combining these modalities may yield a much richer representation of the query with less ambiguity and potentially lead to a better open-set localization performance. \textcolor{black}{In this work, in addition to a sketch query, we use a linguistic definition of the category also known as gloss, e.g. ``a small portable computer small enough to use in your lap" as multimodal queries for the object localization task.}

There are several technical challenges associated with multimodal query-guided object localization, such as (a) a large domain gap between the query modalities (e.g., text, sketches, etc.) and the target natural images, and (b) diverse and minimal information present in queries. For example, a sketch query captures abstract shape information of an object, whereas a text query often captures partial semantic information about the object category. \textcolor{black}{In order to address these challenges, one plausible solution is to use the standard region proposal network (RPN) and score the generated proposals against the query. However, the standard RPN does not utilize query information to generate region proposals; therefore, the relevant region proposals may not even be generated, especially for the open set case. On the contrary, in our framework, we propose a cross-modal attention scheme towards generating object proposals relevant to the input queries. A preliminary version of the same was proposed in our earlier conference paper~\citep{tripathi2020sketch}. The novel extended version of the cross-modal attention strategy is designed to generate a spatial compatibility matrix by comparing the combined query, i.e., concatenated sketch and text representation, with the local image representations obtained from each image feature map location, thereby incorporating query information during proposal generation. In other words, our proposal generation step is query-aware.} A unique advantage of this strategy is that it enables the generation of proposals, even for those object categories that are unseen during training.  
Further, we propose a novel multimodal proposal scoring scheme to score object proposals with features from multiple modalities. The proposed scheme first defines a subspace constructed using queries as the basis vectors, then the feature vector of each proposal is projected onto this subspace. Finally, the projected vector is utilized to score each of the proposals. By being an orthogonal projection, the proposed scheme generates a vector in the subspace of the queries which is closest to the object proposal vector, and hence it leads to better scoring between the queries and proposals. Moreover, the proposed scoring scheme is able to capture complementary information present in multiple modalities that enables it to achieve superior performance for open-set object localization.
 
We have performed extensive experiments on multimodal query-guided object localization on public benchmarks. 
We show results for both the open-set, i.e., disjoint train-test categories, and the closed-set, i.e., common train-test categories settings, and perform extensive ablation studies. 
Our method with sketch and gloss as composed queries achieves 33\% and 10\% mAP for closed-set and open-set object localization, respectively, and significantly outperforms all related baselines. 

Contributions of this paper are listed as follows:
\begin{enumerate}[(I)]
 \item We present an object proposal generation module that is guided by multimodal queries. It proposes a novel extension of our cross-modal attention scheme to generate a spatial compatibility matrix between the different query feature vectors and image features. Being query-aware, this module is capable of generating proposals even for those object categories that are unseen during training. 

 \item We propose a novel orthogonal-projection based proposal scoring scheme that can efficiently score queries from multiple modalities with the object proposals in a better way.

 \item We demonstrate query-guided proposal generation and, finally, instance-level object localization on natural images using the query representation across modalities. Despite the large domain gap between the query (text and sketch) and the target (natural image) data points, we achieve impressive localization performance on challenging public benchmarks. Our method shows impressive performance gain ($\approx 4.7\%$) on open-set object localization.

 
\end{enumerate}

The rest of the paper is organized as follows. Section 2 discusses the existing work from the computer vision literature that is related to this proposed research. Section 3 presents our proposed cross-modal object localization framework, including a detailed analysis of the novel orthogonal projection-based proposal scoring and the cross-modal attention scheme involving multiple query modalities. Section 4 demonstrates the effectiveness of our approach via performing extensive experiments using several publicly available datasets for various modalities, followed by a conclusion in Section 5.

\section{Related work}
\label{sec: relwork}
\subsection{Sketch for vision tasks}
A better understanding of hand-drawn sketches and their utility to computer vision and cognitive science at large has been an active area of research. In order to achieve this goal, developing techniques for a robust representation of sketches has gained huge attention over the last decade. 
In addition to convolutional neural networks~\citep{yu2017sketch}, which are traditionally used, there have been some works that utilize RNN~\citep{ha2017neural} and transformers~\citep{xu2019multi} for learning sketch encoders. 

The area that has significantly benefited from sketch representation techniques is sketch-based image retrieval or SBIR. The goal of SBIR is to retrieve natural images using sketches as queries. Traditional SBIR methods utilize a separate feature computation step that uses manually-tuned features, such as SIFT or histogram of gradients, followed by a bag-of-words encoding as sketch representation~\citep{eitz2010sketch,hu2013performance} and sometimes image edges or contours are also extracted for building image features~\citep{wang2015sketch,DBLP:journals/tmm/ZhangQTHT16}. On the other hand, modern methods leverage deep networks for learning a joint embedding space where sketches and natural images are projected. In these works, often ranking loss such as the contrastive~\citep{sangkloy2016sketchy} or the triplet loss~\citep{yu2016sketch} is used to learn a ranking function between the sketch queries and the candidate images. In~\citep{song2017deep}, researchers have leveraged an attention model to solve fine-grained SBIR and have also introduced higher-order learnable energy function-based loss to alleviate the domain gap between the images and the sketches. \textcolor{black}{ In~\citep{Bhunia2022SketchingWW,Chowdhury2022PartiallyDI}, researchers have tackled the task of noise-tolerant image retrieval.} To improve the efficiency for large-scale image retrieval, hashing models have been explored~\citep{liu2017deep,shen2018zero,xu2018sketchmate,zhang2018generative}. 


Sketches have also been used to study the perceptual grouping ability of machines~\citep{li2018universal,qi2015making} and sketch synthesis~\citep{ge2020creative,ha2017neural,song2020beziersketch}. In our earlier work~\citep{tripathi2020sketch}, we have shown the utility of hand-drawn sketches for object localization in natural images. Although sketches provide critical visual cues, they often lack semantics. To fill this gap, in this work, we propose a method to leverage semantics (using object category name or gloss) along with sketches for object localization.

\subsection{Visual Grounding}
Visual grounding~\citep{liu2020learning,plummer2018conditional,wang2018learning,DBLP:journals/tmm/LiJ18} is a task that has some similarities with the task presented in this paper. However, there are two key differences: (i) visual grounding often restricts itself to natural language query alone, whereas our model supports sketch, object category, and gloss as queries.
(ii) The natural language query in visual grounding describes the object, its attributes, and its relationships with other objects in the image, and it is not an object definition (or gloss) like ours. Further, unlike visual grounding, which leverages large-scale image-caption pairs during training, we only have very few unique definitions (or gloss) for every object and a large number of hand-drawn sketches for training our object localization framework.

\subsection{Object Detection}
Object detection is a core computer vision task. 
Modern object detection methods can be grouped into the following two categories:  
(i) proposal-free methods ~\citep{kong2019foveabox,lin2017focal,liu2016ssd,redmon2016you,redmon2018yolov3,sermanet2013overfeat, DBLP:journals/cviu/YiWM19} and (ii) proposal-based methods ~\citep{cai2018cascade,girshick2015fast,girshick2014rich,he2017mask,he2015spatial,ren2015faster,Zimmermann2019FasterTO}. Proposal-free methods are single-stage detectors, and therefore, they are faster during inference. However, they often fall short of performance as compared to proposal-based approaches. 

Under proposal-based approaches, Girshick et al.~\citep{girshick2014rich} have proposed a two-stage object detection method. In their first stage, they leverage selective search~\citep{uijlings2013selective} to generate object proposals. In the second stage, these generated proposals were classified as one of the object categories using an independently-trained classifier. Ren et al.~\citep{ren2015faster} proposed an end-to-end trainable object detector popularly known as Faster R-CNN. These object detectors are reasonably successful in the closed-set setting. However, they do not generalize well in an open set setting where an object category may or may not be seen during the training phase. 
Recently, Hsieh et al.~\citep{hsieh2019one} have proposed one-shot object detection. In their work, an object image is used as a query, and all the instances of the query object in the target image are detected. However, unlike their work, where query and target images are from the same distribution, i.e., natural images, our queries, i.e., gloss or hand-drawn sketches, are from a significantly different domain than those of the target images.

Object detection in the zero-shot setting has also been studied in the literature~\citep{bansal2018zero,rahman2018zero,rahman2018polarity}. Typically by alleviating the confusion between the ``background" and unseen class, these methods improve object proposal generation for unseen object categories~\citep{bansal2018zero}. Recently, there has been extensive research in context-aware zero-shot detection~\citep{yang2018graph,chen2018context,hu2018relation,luo2020context} which incorporates joint detection of multiple objects~\citep{chen2018context,hu2018relation} or a background scene graph as a knowledge source~\citep{luo2020context}. These works are similar in spirit to the proposed work, but the proposed work is query-guided and utilizes a multimodal query to perform object localization.
 

\subsection{Attention Schemes in Deep Learning Literature}
The use of attention models is prevalent in deep learning literature. They allow the relevant features to become more crucial. Here, we briefly review the utilization of attention in object localization literature.  Choe et al.~\citep{choe2019attention} presented an attention network to score object proposals and showed its utility in object localization. Li et al.~\citep{DBLP:journals/tmm/LiWLDXFY17} proposed Attention to context Convolution Neural Networks (AC-CNN) in object detection to integrate local and global context. Leveraging the self-attention mechanism~\citep{wang2018non},  Heish et al.~\citep{hsieh2019one} presented Co-attention and co-excitation network (CoAtEx) for one-shot object localization. In their work, the response at each feature map location of an image is computed as a weighted combination of the feature vectors at each feature map location of the query. Here weights depend on the similarity between target and query image pixel pairs. It should be noted that both query and target images are from the same modality in CoAtEx. In comparison, proposed cross-modal attention determines the spatial compatibility between global query representation and localized image region representations, thereby mitigating the domain misalignment. In more recent work, authors~\citep{yan2019meta} used class-specific attentive vectors inferred from images of objects in a meta-set to apply channel-wise soft attention to proposals' feature maps. The channel-wise soft attention may not be trivially utilized in our problem setup due to the domain gap between the target and the query.

\subsection{Multimodal learning for Vision}
The natural environment of any visual task contains multiple modalities.  Leveraging data from multiple modalities has been an expanding area of research in the vision literature. Multimodal learning has a very broad range of applications in computer vision, including but not limited to medical image analysis~\citep{james2014medical}, audio-visual speech recognition~\citep{potamianos2003recent}, multimedia event detection~\citep{lan2014multimedia}, multimodal emotion recognition~\citep{soleymani2011multimodal} and visual question answering~\citep{Osman2019DRAUDR}. A key challenge in this area is to summarize information from multiple modalities in a way that is lossless and exploits their complementary or supplementary nature.
In~\citep{calder2001principal,glodek2011multiple}, the authors study the problem of emotion recognition by utilizing facial expressions, head gestures, and other visual cues. Researchers in~\citep{rajagopalan2016extending,ren2016look,vinyals2015show,zadeh2018memory} utilize multi-view LSTM to model cross-view interactions over time or structured data. In the area of image retrieval, composing multimodal queries has gained interest in the last few years~\citep{tirg}. In~\citep{dang2012supervised}, researchers incorporate semantic and geographical information to improve image retrieval, while researchers in~\citep{cao2014medical} utilize both textual and visual features for improved image retrieval performance in the medical domain. To fuse multimodal input in information retrieval, concatenation~\citep{pham2007latent} of features and multi-layer probabilistic latent semantic analysis (PLSA)~\citep{hofmann2013probabilistic} models~\citep{cao2014medical} have been proposed.  Attribute as operator~\citep{nagarajan2018attributes}, and parameter hashing~\citep{noh2016image} methods create a transformation matrix from text and use it to transform the image features. Researchers in~\citep{zadeh2016mosi,morency2011towards,rosas2013multimodal} utilized visual cues for sentiment analysis in product and movie reviews, which conventionally used only text. They directly concatenated visual and textual representations in order to obtain a joint representation. Tensor fusion network~\citep{zadeh2017tensor} was proposed to fuse up to three different modalities for a multimodal sentiment analysis task. \textcolor{black}{More recently, authors in~\citep{arsha2021attbot} proposed attention bottlenecks in transformers to effectively fuse features from videos and audio, and in ~\citep{akbari2021vatt} three separate transformer models were trained using self-supervised learning with multi-modal contrastive losses to extract effective multi-modal representation from raw inputs of video, audio, and text.} The reader is encouraged to read elaborate surveys\textcolor{black}{~\citep{UPPAL2022149,baltruvsaitis2018multimodal}} on multimodal learning to know more about this area. Our work is closely related to this line of study, where we use cues from a natural image, a sketch, and text to perform object localization. 

\section{Our Approach}
\label{sec:model}
In this section, we first provide a formal introduction to the multimodal query-guided object localization problem. Then, we present our solution by first describing the cross-modal attention scheme for either of the query modalities. We consider the specific examples of text and hand-drawn sketches as the query modalities in this paper and show how our cross-modal attention can be leveraged for each modality to generate proposals relevant to the query objects. We then introduce our novel orthogonal projection-based proposal scoring scheme to better score object proposals with respect to queries of multiple modalities. 

\begin{figure*}[h]
    \centering
    \includegraphics[width=\textwidth]{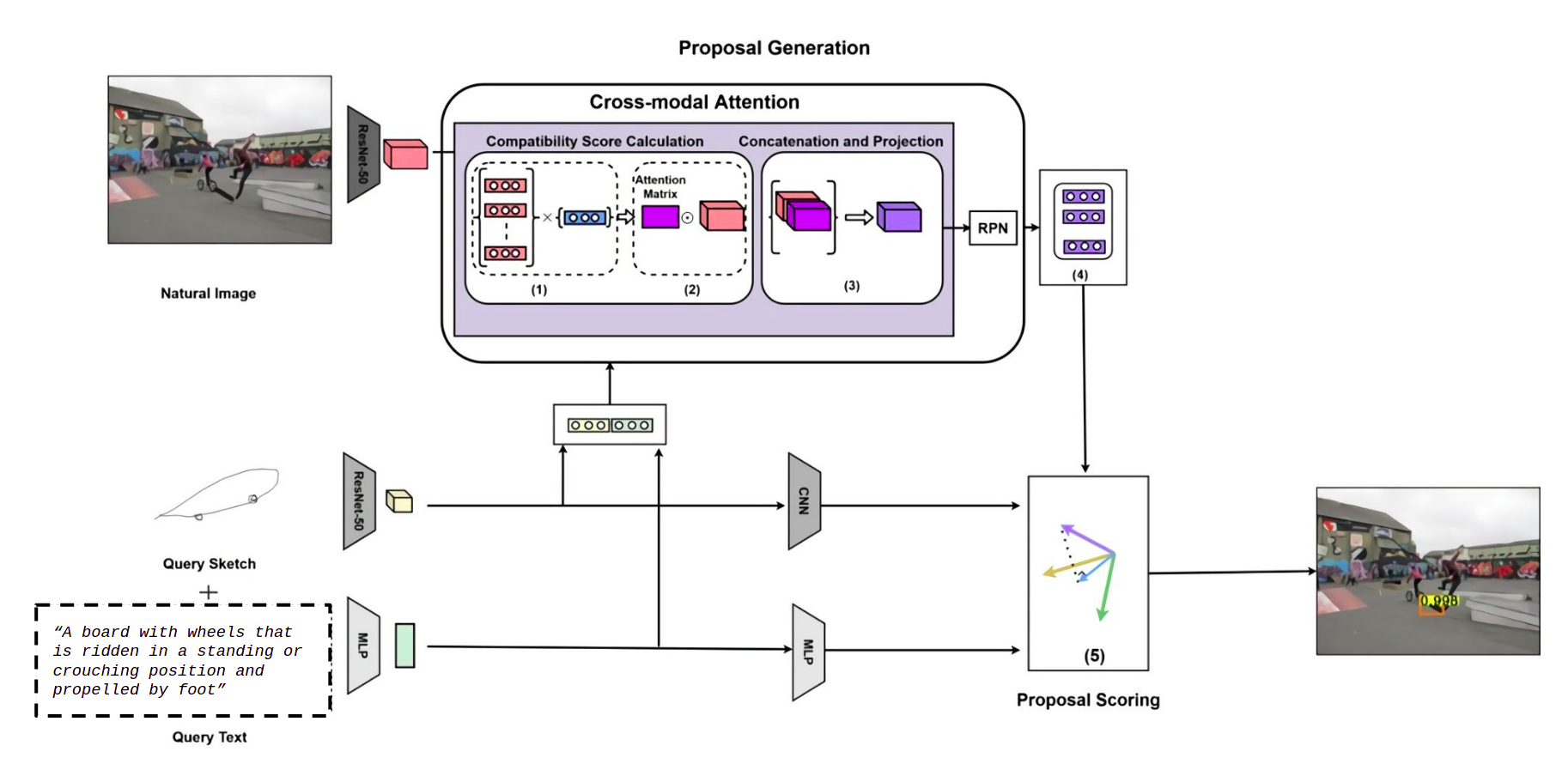}
    \caption{\label{fig:approach} Given an image and queries of different modalities, our object localization framework works in the following two stages: (i) \textbf{query-guided proposal generation:} in this step, the global fused feature vector of different queries that are shown using blue color is scored with the image feature vectors that corresponds to each location on the image feature map that is shown using pink color to generate the spatial compatibility also called the attention scores. (Block 1). Next, these attention scores, which are shown using violet color, are multiplied with the image feature maps, which are shown using pink color to get the attention features (Block 2). Before passing it through the region proposal network (RPN), it is first concatenated with the original feature maps and projected to the original dimension. The RPN is able to generate relevant object proposals because of the spatial compatibility, that is integrated into the image feature maps, between global fused queries representation and regional image representation (Block-3), (ii) \textbf{orthogonal-projection based proposal scoring:} the representation for each of the pooled object proposals that are shown using indigo is scored with query feature vectors from multiple modalities to generate localization for the object of interest (Block-5). The proposal vector is projected onto the subspace spanned by the queries, and the projection vector is utilized to query against the proposal vector. \textbf{[Best viewed in color].}}
\end{figure*}
\subsection{Problem Formulation}
Let $I = I_{train} \bigcup I_{test}$ be a set of all-natural scene images in a dataset ${{\cal D}_I}$, each containing a variable number of object instances and categories. Here $I_{train}$ and $I_{test}$ are sets of train and test images, respectively. Like any other machine learning task, these two sets are mutually exclusive, and only $I_{train}$ is available during training.  Further, let $S=S_{train}\bigcup S_{test}$ be a set of all sketches, each containing one object,  $T=T_{train}\bigcup T_{test}$ be the set of a textual description of an object (either object category name or gloss of the object category), and $C =C_{train}\bigcup C_{test}$ be a set of all object categories. During training, each training sample contains an image $i \in I_{train}$, a sketch query $s_c \in S_{train}$, a text query $t_c \in T_{train}$ where $c \in C_{train}$, and all the bounding boxes corresponding to object category $c$ in the image $i$. At test time, given an image
$i^{'} \in I_{test}$ and a sketch query $s_{c'} \in S_{test}$, a text query $t_{c'} \in T_{test}$, where $c^{'} \in C_{test}$, the problem is to localize all the instances of the object category $c^{'}$ in the image $i^{'}$. Note that we show experimental results in cases where $C_{train} = C_{test}$, i.e., categories in $C_{test}$ are seen during training time ({closed-set object localization}), as well as $C_{train} \bigcap C_{test} = \phi$, i.e., categories in $C_{test}$ are not seen during training ({open-set object localization}). 

The proposed multimodal query-guided object localization is end-to-end trainable, and it works in the following two stages: (i) query-guided object proposal generation (Section~\ref{sec:attention}), and (ii) orthogonal-projection based proposal scoring (Section~\ref{sec:scoring}). Fig.~\ref{fig:approach} shows a schematic diagram of the proposed framework.

\subsection{Cross-modal Attention for Query-guided Object Proposal Generation}
\label{sec:attention}
Faster R-CNN~\citep{ren2015faster} is a popular framework for two-stage object detection, and in the first stage, it uses a region proposal network (RPN) to generate object proposals. A vanilla region proposal network could be used to generate object proposals in our task. However, a traditional RPN is not built to take advantage of any query-level information on object appearance or semantics. As a result, the object proposals that are relevant to the sketch or textual queries may not even be generated, particularly when the object of interest is of low resolution, occluded, hidden among other objects that are better represented in the input images, or most importantly, is one of the categories which is unseen during training. Therefore, using an RPN in its vanilla form may not suffice in our pipeline. To address the aforementioned problem, we proposed cross-modal attention to incorporate the sketch query information in the RPN in our earlier work~\citep{tripathi2020sketch}. In this work, we adapt the cross-modal attention to incorporate multimodal queries in the RPN and thereby guide the proposal generation. Regions of interest (ROIs) are pooled from region proposals generated using RPN utilizing a strategy similar to the Faster R-CNN, and a scoring function $\Theta$ is learned between these ROIs and joint representations of sketch and text queries.

We now describe our cross-modal attention framework to generate object proposals relevant to the queries of different modalities. A preliminary version of this, specific to sketch queries only, was presented in our earlier work~\citep{tripathi2020sketch}. In this work, we extend the framework to include additional modalities, such as text queries.  We feed a joint representation of sketch and text modalities to the proposal generation module, which is trained to produce a spatial weight map that provides high scores to the areas on the target image which are visually or semantically similar to the object corresponding to the given query(ies).

As mentioned earlier in this paper, we consider the examples of two query modalities, i.e., sketch and text. Suppose a sketch $s_c \in S$ and a text $t_c \in T$ (either category name or gloss) of an object category $c \in C$ is used to query an image $i\in I$. To generate the feature representation of images and sketches, we use ResNet-50 models pretrained on Imagenet~\citep{deng2009imagenet} and Quick Draw~\citep{jongejan2016quick} datasets, respectively, as backbones. We use either of the two types of text queries: object category name and generic object description (aka gloss). Feature representations for these are obtained using a language encoding scheme followed by a trainable multi-layer neural network. The de facto choice for language encoding now is fine-tuned BERT~\citep{devlin2018bert} model. We used them to represent the object category name and its gloss, respectively. 
Suppose $\phi_I$, $\phi_T$, and $\phi_S$ represent these backbone feature encoders, then image, text, and sketch feature maps are computed as: 

\begin{equation}
    i^{\phi_I} = \phi_I(i)\quad \text{,}\quad s^{\phi_S}_c = \phi_S(s_c)\quad \text{and} \quad t_c^{\phi_T} = \phi_T(t_c),
\end{equation}
where, $i^{\phi_I} \in \mathbb{R}^{w\times h \times d}$, $t^{\phi_T} \in \mathbb{R}^d$, and $s^{\phi_S}_c \in \mathbb{R}^{w'\times h' \times d}$ are the extracted image, text, and sketch feature representations respectively.
From these feature maps, the compatibility score is learned between the sketch and the text queries, and the image feature maps by first applying non-linear transformations as below:
\begin{equation}
    i^{\psi_I} = \psi_I(i^{\phi_I})\,\, \text{,}\,\, s_{c}^{\psi_S} = \psi_S(s_{c}^{\phi_S})\quad \text{and} \quad t_c^{\psi_T}=\psi_T(t_c^{\phi_T}).
\end{equation}

A set of local feature vectors is formed by obtaining one vector at each location $(m,n)$ in the image feature map $i^{\psi_I}$, where $m\in \{1, 2, \dots, w\}$ and $n\in \{1,2,\dots, h\}$. Each vector represents a spatial region on the target image, and the set gives us the spatial distribution of the features. Subsequently, this is compared against a fusion of global representation of the sketch features and textual features. For image feature map i.e. $i^{\psi_I} \in \mathbb{R}^{w\times h \times d}$, the extracted set of feature vectors is represented as $L^i= \{\mathbf{L}_1^i, \mathbf{L}_2^i,...,\mathbf{L}_{w \times h}^i \}$ where $\mathbf{L}_j^i \in \mathbb{R}^{1\times 1\times d}$ $\forall$ $j\in \{1,2,\dots, w\times h\}$.

In the case of sketches, a global representation of sketch feature maps is obtained via the global max pool (${\cal GMP}$) operation, i.e., $\mathbf{L}_g^{s_c} = {\cal GMP} (s_c^{\psi_S}),$,
where, $\mathbf{L}_g^{s_c} \in \mathbb{R}^{1\times 1 \times d}$. The sketch and text representations are concatenated and projected to obtain the final query representation.
\begin{equation}
    \mathbf{L}_g^{q_c} = W[(\mathbf{L}_g^{s_c})^T; t_c^{\psi_T}],
\end{equation}
where, $W \in \mathbb{R}^{d\times 2d}$. 
A spatial compatibility score between $\mathbf{L}_j^i \in L^i$ and $\mathbf{L}_g^{q_c}$, is computed as follows:
\begin{equation}
\label{eq:constant}
    \lambda(\mathbf{L}^i_j, \mathbf{L}_g^{q_c}) = \frac{\mathbf{L}_j^i \cdot \mathbf{L}_g^{q_c} }{\cal K} \,\,\,\,\, ,
\end{equation}

where ${\cal K}$ is a constant. For simplicity of notation, we will refer to the left-hand side of Equation~(\ref{eq:constant}) as $\lambda_{jg}$ from here onwards.

It should be noted that these compatibility scores are generated as a spatial map, which can be understood as a 2D-map representing attention weights. Therefore, in order to obtain attended feature maps, we perform element-wise multiplication of these compatibility scores and the original image feature map at each spatial location, i.e.,
\begin{equation}
    i^{a_I}_j = i_{j}^{\phi_I}\odot \lambda_{jg}, 
\forall j \in \{1,2,\cdots,w \times h\}.
\end{equation}
This attention feature map aims to capture information about the location of objects in an image that shares high compatibility score with both the sketch query and the text query. Therefore, to incorporate this information, attention feature maps are concatenated along the depth with the original feature maps, i.e., $i^{\phi_I}_f = [(i^{a_I})^T;(i^{\phi_I})^T]^T$, where $i_f^{\phi_I}\in \mathbb{R}^{w\times h \times 2d}$. These concatenated feature maps are projected to a lower-dimensional space to obtain the final feature maps, which are subsequently passed through the RPN to generate object proposals relevant to the sketch query. 

\subsection{Orthogonal-projection based Proposal Scoring}
\label{sec:scoring}
Once a small set of proposals represented as $R_i$ for $i\in I$ are pooled from all query-guided region proposals generated by the RPN, feature vectors for these proposals are computed along with the final feature vectors for sketch and text query, respectively, as follows. 
\begin{equation}
    r_k^{\phi'_I} = \phi'_I(r_k^{\phi_I})\,\,\text{,}\,\, s_c^{\phi'_S} = \phi'_S(s_c^{\phi_S})\,\, \text{and}\,\, t_c^{\phi'_T} = \phi'_T(t_c^{\phi_T}),
\end{equation}
where $r_k^{\phi'_I} \in \mathbb{R}^d$ is generated using standard Faster R-CNN protocols, $s_c^{\phi'_S}\in \mathbb{R}^d$, $t_c^{\phi'_T}\in \mathbb{R}^d$, $r_k \in R_i$. The $\phi'_I$ and $\phi'_S$ are two separate multi-layer CNN followed by mean pool, and $\phi'_T$ is the multi-layer feed-forward neural network. In order to rank these object proposals with respect to the multimodal query representation, a scoring function $\Theta$ is learned. During training, the region proposals are labeled as foreground (or 1) when they have $\geq 0.5$ intersection over union (IoU) with the ground truth bounding boxes and the objects in the bounding boxes belong to the same class as the query, or else they are labeled as background (or 0). Then, we minimize a margin rank loss between the representations of the generated object proposals and the queries such that object proposals that contain the object of the same class as the queries are ranked higher.

Object proposals belong to a domain different from the queries, which themselves are from entirely different domains. Further, queries from different domains may capture different kinds of information, e.g., the sketch of an object captures the shape information, while on the other hand, text captures the semantics of the object. Therefore, we need to compare the object proposals against both these kinds of information to obtain a better score. In order to ensure better scoring, we propose orthogonal-projection-based proposal scoring. We begin by finding the proposal feature vector's projection in the subspace defined by the queries. We then use that projection to compute a score with the representation of the proposal. By being an orthogonal projection, we use the closest vector containing the complementary information present in the sketch and the text, in the query space, to the representation of the proposal for proposal scoring. We now describe our proposed orthogonal projection scheme in detail.

\subsection{Orthogonal Projection:}
We construct a vector subspace $\mathcal{M}$ by considering the queries $s^{\phi'_S}_c$ and $t^{\phi'_T}_c$ as the basis vectors. Then, we perform an orthogonal projection of the object proposal vectors into this subspace. This projection yields a  vector that contains the complementary information present in $s^{\phi'_S}_c$ and $t^{\phi'_T}_c$ and is closest to the proposal vector. To obtain the orthogonal projection, we first define a matrix $B_c = [s^{\phi'_{S}}_c, t^{\phi'_T}_c] \in \mathbb{R}^{d\times 2}$, and the projection matrix is defined in terms of $B_c$ as follows:
\begin{equation}
P_{R(\mathcal{M})} = B_c(B^T_c B_c)^{-1}B^{T}_c,    
\end{equation}
where $P_{R(\mathcal{M})}$ is the projection matrix on the range space of $\mathcal{M}$ i.e. $R(\mathcal{M})$. In order to obtain the fusion, we project $r_k$ onto the $R(\mathcal{M})$, i.e.
\begin{equation}
    q_k^c = P_{R(\mathcal{M})} r_k^{\phi'_I},
\end{equation}
where $q_k^c \in \mathbb{R}^d$ is the fused sketch and text feature vector corresponding to the object proposal $r_k$.

In order to learn the scoring function $\Theta$, the object proposal feature vectors are concatenated with the feature vector obtained before. These concatenated feature vectors are passed through the scoring function (a one-layer neural network in our framework), and it predicts the foreground probabilities of the proposals with respect to the fused query. Let $a_k$ be the predicted foreground probability for proposal $r_k \in R_i$, and it is given by $a_k = \Theta([(r_k^{\phi'_I})^T;(q_k^c)^T]^T)$,
where, both $r_k^{\phi'_I}$ and  $q_k^c$ are defined in Section~\ref{sec:scoring}. Now, towards training the scoring function $\Theta$, a label $y_k = 1$ or $0$ is assigned to $r_k$ depending on its overlap with a ground truth object bounding box, as defined in the previous paragraph. Instead of using a neural network, cosine similarity can also be used to compute the score.
Motivated from~\citep{hsieh2019one}, the loss function used in training is defined as:

\begin{multline}
\label{eq:ce}
    L(R_i, s_c) = \sum_{k}\big\lbrace y_k  \max(m^+-a_k,0) +
    (1-y_k) \max(a_k-m^-,0) + L_{MR}^k\big\rbrace
\end{multline}
\begin{equation}
\label{eq:margin}
L_{MR}^k =\sum_{l=k+1}\big\lbrace\mathbf{1}_{[y_l=y_k]} \max(\mid a_k-a_l\mid -m^-,0) 
+ \mathbf{1}_{[y_l\neq y_k]} \max(m^+-\mid a_k-a_l\mid,0)\big\rbrace,    
\end{equation}

where $m^+$ and $m^-$ are positive and negative margins, respectively. The above loss function consists of two parts: (i) In Equation (\ref{eq:ce}), the first part of the loss function assures that the object proposals that are overlapping with the ground truth object locations are predicted as foreground with high probability. (ii) The second part of the loss function, i.e., Equation (\ref{eq:margin}), is a margin-ranking loss that takes pairs of the proposals as input. It aids in reinforcing a greater division between prediction probabilities of foreground and background object proposals, and therefore, it improves the ranking of all the foreground proposals overlapping with the true location(s) of the object of interest. \textcolor{black}{Both parts of this loss function in equation~\ref{eq:ce} are equally weighted during training}. Additionally, a cross-entropy loss on the labeled (background or foreground) feature vectors of the region proposals and a regression loss on the predicted bounding box location deltas (same regression loss as in Faster-RCNN) with respect to the ground truth bounding box are used for training.

Moreover, using the orthogonal projection scheme described before can also be viewed as a fusion technique that can fuse a number of queries of multiple modalities without requiring any additional parameters. An important objective of a fusion technique is that the resultant fused representation has better utility than the individual queries. This property can be meaningfully encoded as the following equations: 
\begin{equation}
\label{eq:dist_eq_1}
    d(r_k^{\phi'_I}, f(s_c^{\phi'_S}, t_c^{\phi'_T})) \leq d(r_k^{\phi'_I}, s_c^{\phi'_S}) 
\end{equation}

\begin{equation}
\label{eq:dist_eq_2}
     d(r_k^{\phi'_I}, f(s_c^{\phi'_S}, t_c^{\phi'_T})) \leq d(r_k^{\phi'_I}, t_c^{\phi'_T}),
\end{equation}
where $d(\cdot, \cdot)$ is a suitable distance function and $f(\cdot, \cdot)$ is a function that fuses $s_c^{\phi'_S}$ and $t_c^{\phi'_T}$. The objective of enforcing this constraint is that by design, the representations of the query modalities must improve in utility on fusion, i.e., the fused representation should be closer to the feature obtained from an object proposal than any individual query features as measured by a suitable distance function. 
Utilizing the orthogonal projection for scoring inherently enforces the constraints defined in Equation~(\ref{eq:dist_eq_1}) and~(\ref{eq:dist_eq_2}). Therefore, it could be viewed as a fusion technique that utilizes the proposal representation for better scoring without requiring additional parameters.

\section{Experiments and Results}
\label{sec: expts}
\subsection{Datasets}
We evaluate the performance of the proposed framework using the following datasets: 
\subsubsection{\textbf{QuickDraw}~\citep{jongejan2016quick}} It is a large-scale hand-drawn sketch dataset. It contains 50 million hand-drawn sketches of $345$ object categories in all. In our experiments, we selected those sketch categories that overlap with MS-COCO or PASCAL-VOC, as described in the subsequent paragraphs. QuickDraw sketches are stored as vector graphics, and we rasterized the sketches before feeding them into the ResNet.
\subsubsection{\textbf{MS-COCO}~\citep{lin2014microsoft}} It is a de facto natural scene dataset for studying object detection. It contains object bounding box annotations for $80$ object categories. Between MS-COCO and QuickDraw datasets, $56$ object categories are common. Therefore, we randomly selected a total of $800K$ sketches across these common classes for our experiments. The model is trained on the COCO-Train-2017 and evaluated on the MS-COCO-Val-2017 dataset.
\subsubsection{\textbf{PASCAL VOC}~\citep{Everingham10}} It is another common object detection dataset. It contains a total of $20$ object classes. We choose images of nine object categories that are common to the QuickDraw dataset for our experiments. We trained our model on the union of VOC2007 train-val and VOC2012 train-val sets and evaluated on the VOC-test-2007 set.

\subsubsection{\textbf{Gloss dataset}} Semantic information about the object is introduced to our localization framework by utilizing an embedding of a brief sentence describing an object category, also known as gloss. We collected a gloss of object categories selected from the Visual Genome~\citep{krishna2017visual} and MS-COCO datasets from WordNet~\citep{fellbaum2012wordnet}. We refer to this collection as the Gloss dataset. This dataset contains gloss for \textit{1615} object categories in all. Some examples of this dataset include gloss for sofa is \emph{an upholstered seat for more than one person}, gloss for carrot is \emph{deep orange edible root of the cultivated carrot plant}.

\begin{table*}[!t]
\renewcommand{\arraystretch}{1.0}
        \centering
        \begin{tabular}{lccccc}
        \toprule
        \multirow{2}{*}{Method}& \multirow{2}{*}{Fusion} & \multicolumn{2}{c}{Open Set} & \multicolumn{2}{c}{Closed Set} \\
        && \%AP@50 &  \%mAP& \%AP@50 &  \%mAP       \\ 
        \midrule
        Modified Faster RCNN&-&7.4&5.4&31.5&18.0\\
        CoATex~\citep{hsieh2019one}&-&12.4&6.3&48.5&28.0\\
        Ours & & &\\
        ~~~~Sketch only~\citep{tripathi2020sketch}&-&15.0&7.4 &50.0&30.1\\
        ~~~~Gloss only &-& 15.2 & 7.6& 54.2 & 32.7\\
        ~~~~Sketch + gloss & Late &16.0&7.8 &53.3&32.5\\
        ~~~~Sketch + gloss & Concat &18.8&9.6 &53.4&32.6\\
        ~~~~Sketch + gloss & OPS &\textbf{19.7}&\textbf{10.0}&\textbf{54.4}&\textbf{33.0}\\
         \bottomrule
        \end{tabular}%
        \caption{\label{tab:multi_oneshot} 
       \textit{Results in one-shot open-set and closed-set settings on the MS-COCO-Val-2017 dataset.}  Bringing semantics using the additional queries, such as gloss and object category names, generally has a positive effect on the localization performance. Further, orthogonal projection-based scoring. clearly outperforms other fusion techniques in challenging open-set settings. }
\end{table*}

\subsection{Baselines and Our Variants}
In order to demonstrate the superior performance of our approach, we adapt and compare it with the following popular approaches from the object detection and image-guided localization literature:
\subsubsection{\textbf{Sketch-only baselines}} In this section, we describe the baselines to evaluate the sketch-only object localization.

\noindent{\textbf{Modified Faster R-CNN~\citep{ren2015faster}}:} For query-guided object localization tasks with a sketch query, we adapt Faster RCNN. Towards this end, during training, if an object instance in an image belongs to the same class as the sketch query, we assign class label $1$ to it and $0$ otherwise. We then generate object proposals using the vanilla region proposal network (RPN). Each region proposal is then identified as background or foreground using a binary classifier. To this end, the region of interest features for each region proposal is first concatenated with the query features and then passed through the binary classifier. We also used a triplet loss to rank the object region proposals concerning the sketch query. This baseline is referred to as modified Faster R-CNN in this paper. 
\newline
\noindent{\textbf{Co-attention and co-excitation network (CoATex)~\citep{hsieh2019one}}:} It is a recent one-shot object localization method using image queries. The query information is integrated into the image feature maps by utilizing non-local neural networks~\citep{wang2018non} and channel co-excitation~\citep{hu2018squeeze}. This method is adapted to work with the sketch query directly, and it is used as a second baseline in our experiments.

The feature extractors for images and sketches are ResNet-50 models pre-trained on Imagenet and QuickDraw, respectively, for both these baseline methods.

\subsubsection{\textbf{Our variants}}
\textcolor{black}{In order to perform a comparative study with the  above-mentioned baseline approaches, we present the following variants of our approach:}
\newline\noindent\textbf{Sketch only~\citep{tripathi2020sketch}:} \textcolor{black}{In this variant of our approach, we only use sketch queries to localize objects in our framework. }
\newline
\noindent\textbf{Gloss only:}\textcolor{black}{In this variant of our approach, we only use natural language description (aka gloss) of object categories to localize them in a natural scene. This variant is useful to demonstrate the effectiveness of our approach in cases there is no visual query available. }
\newline
\noindent\textbf{Sketch+Gloss:} \textcolor{black}{This is our full model. In this, we fuse two modalities, namely visual (sketch) and textual (gloss), using the following different fusion strategies.}  
(i) \textbf{Late fusion:}
Let the sets $R^s$ and $R^t$ be the set of proposals obtained after comparing with $s_c$ and $t_c$ respectively at the test time, where $s_c$ and $t_c$ are sketches and text queries of class $c\in C$ respectively. In late fusion, we take the union of these sets of proposals and choose the Top-$N$ proposals as the final set. (ii) \textbf{Concatenation fusion:}
Let $s_c$ and $t_c$ be the sketch and text queries, respectively. In this fusion strategy, these queries are concatenated and projected to obtain the fused query. (iii) \textbf{Proposed OPS as fusion:} Finally, we use the proposed orthogonal projection scoring (OPS) presented in Section~\ref{sec:scoring} to fuse sketch and gloss embeddings. 



\begin{table}[]
\centering
\begin{tabular}{lc}
\toprule
Method              & mAP  \\
\midrule
Modified Faster RCNN & 0.65 \\
CoATex~\citep{hsieh2019one}               & 0.61 \\
Ours (Sketch only)~\citep{tripathi2020sketch}   & 0.65\\
\bottomrule
\end{tabular}
\caption{\label{tab:voc_full}
    \textit{Results in one-shot open-set setting on VOC test-2007 dataset.}  }
\end{table}


\subsection{Experimental Setup}
\label{sec:setup}
In order to get the feature representation for the images and the sketches, we used two ResNet-50 models pre-trained on Imagenet~\citep{deng2009imagenet} and a subset of 5 million images from QuickDraw~\citep{jongejan2016quick}, respectively. We use hand-drawn sketches from the common classes of QuickDraw to localize objects in images from MS-COCO and PASCAL-VOC datasets. Once the gloss dataset is created, we use WordNet synset matching to retrieve a set of similar object categories for each class. This set of similar categories is utilized to fine-tune a pretrained BERT model~\citep{devlin2018bert} under the objective that similar classes' representation is close to each other than the non-similar classes. We evaluate the performance of our model under closed-set and open-set settings.

\subsubsection{Open-set experimental setting} 
In the open-set experimental setting, out of the 56 common classes across COCO and QuickDraw, we choose  42 and 14 classes as `seen' and `unseen' categories, respectively. The `seen' and `unseen' splits are mutually exclusive in terms of object categories and labeled bounding boxes present to ensure the one-shot open-set experimental setting. Our model is trained exclusively on the dataset from the `seen’ classes, and only the `unseen’ classes are used for open-set evaluation. Similarly, for the PASCAL-VOC, out of the nine classes common to QuickDraw, three and six are chosen arbitrarily as `unseen' and `seen'  categories, respectively. The image encoder is pretrained on the Imagenet dataset, except for 14 `unseen' classes and all associated classes obtained by matching their WordNet synsets. Similarly, except for the 14 categories in the `unseen' set, the sketch encoder is pretrained using all of the QuickDraw categories.

\subsubsection{Closed-set experimental setting} In this experimental setting, all the 56 common classes in MS-COCO and Quickdraw datasets are used during training, and the model is evaluated on all 56 categories at test time. Similarly, for the PASCAL-VOC dataset, all data points which correspond to 12 classes, which are common with QuickDraw, are used during the training. During the evaluation, the dataset from all 12 classes is utilized.

\subsection{Implementation Details}
We use PyTorch v1.0.1 with CUDA 10.0 and CUDNN v7.1 to train the model using stochastic gradient descent (SGD) with a momentum of $0.9$ on one NVIDIA 1080-Ti with a batch size of $10$. The learning rate was initially set at $0.01$, but it decays with a rate of $0.1$ after every four epochs, and it is trained for $30$ epochs. The constant $\cal K$ in Equation (\ref{eq:constant}) is fixed at $256$ and $m^+=0.3$ and $m^-=0.7$ in Equation (\ref{eq:ce}) and (\ref{eq:margin}) for all experiments. To obtain the sentence embeddings average of the features of the final layer of BERT is utilized. The BERT model is fine-tuned with a learning rate of $5e-5$, and triplet loss is used during fine-tuning along with the hard-negative mining on the mini-batch. For optimal results, {the cross-modal attention} model is trained incrementally. Firstly, the localization model is trained without attention. Then, the attention model is added to it, and it is trained again. The training protocol is the same as explained before, and it is the same for both steps.


 \begin{table}[!t]
     \renewcommand{\arraystretch}{1.0}
    \centering
    \begin{tabular}{lcc}
    \toprule
    Modality& \%AP@50 & \%mAP        \\ 
    \midrule
    Sketch only& 15.0&7.4\\
    Gloss only& 15.2&7.6\\
    Sketch + gloss& 19.7&10.0\\
    Sketch + gloss with class&20.3&10.6\\
    Sketch + gloss + class&\textbf{23.6} &\textbf{12.9}\\
     \bottomrule
    \end{tabular}%
    \caption{\label{tab:ablation}
    \textit{Results in one-shot open-set setting on MS-COCO-Val-2017 dataset.} This table shows if the category name is available for query, the localization performance of our method can be further improved.} 
    \end{table}

\subsection{Results and Discussion}
\label{sec:multimodal_results}
\begin{table*}[h]
\renewcommand{\arraystretch}{1.0}
            \centering
            \begin{tabular}{lcccc}
             \toprule
        \multirow{2}{*}{Method}& \multicolumn{2}{c}{Open Set} & \multicolumn{2}{c}{Closed Set} \\
        & \%AP@50 &  \%mAP& \%AP@50 &  \%mAP       \\ 
        \midrule
            Ours (Sketch only)~\citep{tripathi2020sketch}~~& 15.0& 7.4&50.0&30.1\\
               ~~+\textcolor{black}{Feature Fusion(3 Sketches)} & \textcolor{black}{17.1}&\textcolor{black}{7.3} &\textcolor{black}{51.9}&\textcolor{black}{31.0}\\
             ~~+Feature Fusion (5 Sketches) & 16.3&7.6 &52.6&32.0\\
             ~~+\textcolor{black}{Attention Fusion(3 Sketches)} & \textcolor{black}{17.6}& \textcolor{black}{7.5}&\textcolor{black}{52.0}&\textcolor{black}{31.0}\\
             ~~+Attention Fusion (5 Sketches) &17.1& 8.0&53.1&32.0\\
             ~~+Gloss& \textbf{19.7}&\textbf{10.0}&\textbf{54.4}&\textbf{33.0}\\
            \bottomrule
            \end{tabular}%
            \caption{\label{tab:few_shot_coco}
            \textit{Comparison of the multi-sketch localization with multimodal localization in multi-query closed-set and open-set categories setting on the MS-COCO-val-2017 dataset. } Here, five sketches mean we use five randomly selected sketch queries. Further, Open Set and Closed Set represent disjoint and common train and test categories, respectively.}
\end{table*}

We now quantitatively and qualitatively evaluate our model in different settings on the MS-COCO and PASCAL-VOC datasets. Our model on the MS-COCO dataset is compared against other related approaches in Table~\ref{tab:multi_oneshot} in both open and closed-set settings. For the sketch-only experiments, the proposed cross-modal attention model significantly outperforms both modified Faster-RCNN and CoATex-based baselines. This is primarily because, unlike faster R-CNN, the cross-modal attention framework effectively incorporates the query information using spatial compatibility (attention) maps to generate region proposals that are relevant. Further, the CoATex baseline~\citep{hsieh2019one} utilizes the non-local feature maps and channel co-excitation module, and these modules are sensitive toward the domain gap present between query and image feature maps in our task. The proposed method, on the other hand, addresses this by computing a spatial compatibility (attention) map directly. Our model, by virtue of cross-modal attention, integrates query information in the image feature map before feeding it through the region proposal network. Consequently, our model is intrinsically able to generate relevant object proposals even for unseen object categories. As a result, our approach outperforms the baselines on unseen object categories.

However, when comparing the sketch-only model with the gloss-only model, the gloss-only model performs significantly better. Both these models have similar architectures aside from the modality of the query. Therefore, it indicates that text-only queries contain information that can be utilized better for object localization. Furthermore, we combine both the sketch query and the gloss query in our method, utilizing orthogonal-projection based scoring, and find that it leads to significant improvement in the localization performance; for example, in the close-set setting, we get 3.8\% improvement when sketch queries are used along with object gloss. This improvement is even more significant (i.e., 4.7\%) in an open-set setting, indicating that incorporating semantic information and shape information helps create a better representation for object localization (Refer Table~\ref{tab:multi_oneshot}).

Compared with the multimodal fusion baselines, the proposed orthogonal-projection-based proposal scoring scheme shows significant performance improvement, indicating that the proposed scheme is able to better capture the complementary information present in multiple modalities. The late fusion technique combines the predicted localization from each of the queries, indicating that combining predictions from multiple sub-optimal queries does not give a sufficient improvement in performance. The concatenation fusion does not impose any constraints on the fusion output and, therefore, leads to sub-optimal results. 

The results for the PASCAL-VOC dataset are reported in Table~\ref{tab:voc_full}. PASCAL-VOC is a small-size dataset, and for our experimental setting, it contains a small number of training images ($\approx$ 9K) with inadequate variability between the classes present in the dataset. It should be noted here that the training set contains only nine classes that are common with the QuickDraw dataset. The proposed method is comparable to the modified faster-RCNN baseline suggesting that query-guided object localization is hard in case of insufficient data.\textcolor{black}{The CoATex baseline suffers degradation in performance, indicating that it is unable to integrate sketch information in the image feature map during proposal generation in the case of small data size and large domain gap. Due small training size, gloss query experiments are not performed on PASCAL-VOC.}

\begin{table}[t]
\centering
\small
\begin{tabular}{lcccc|c}
\toprule
Fusion & \multicolumn{1}{c}{Split1}                          & \multicolumn{1}{c}{Split 2}                         & \multicolumn{1}{c}{Split 3}                         & \multicolumn{1}{c}{Split 4}                         & \multicolumn{1}{|c}{Avg}                             \\

      & \multicolumn{1}{c}{AP@50} &  \multicolumn{1}{c}{AP@50} & \multicolumn{1}{c}{AP@50} &  \multicolumn{1}{c}{AP@50} &  \multicolumn{1}{|c}{AP@50} \\
\midrule
Concat                      & 18.8                                           & 16.4                                           & 18.1                                           & 17.8                                           & 17.8                      \\
OPS                        & \textbf{19.7}                                           & \textbf{17.6}                                           & \textbf{19.9}                                           & \textbf{18.9}                          & \textbf{19.0}                      \\
\bottomrule
        
\end{tabular}
\caption{\label{tab:data_splits} The performance comparison of the proposed OPS scoring with the Concatenation fusion for different sets of classes in the Open set. We used the sketch of the class along with the Gloss of the class in this experiment. The experiments are performed on \textit{MS-COCO} dataset.}
\end{table}

 \begin{figure*}[h!]
    \centering
    \includegraphics[width= \textwidth]{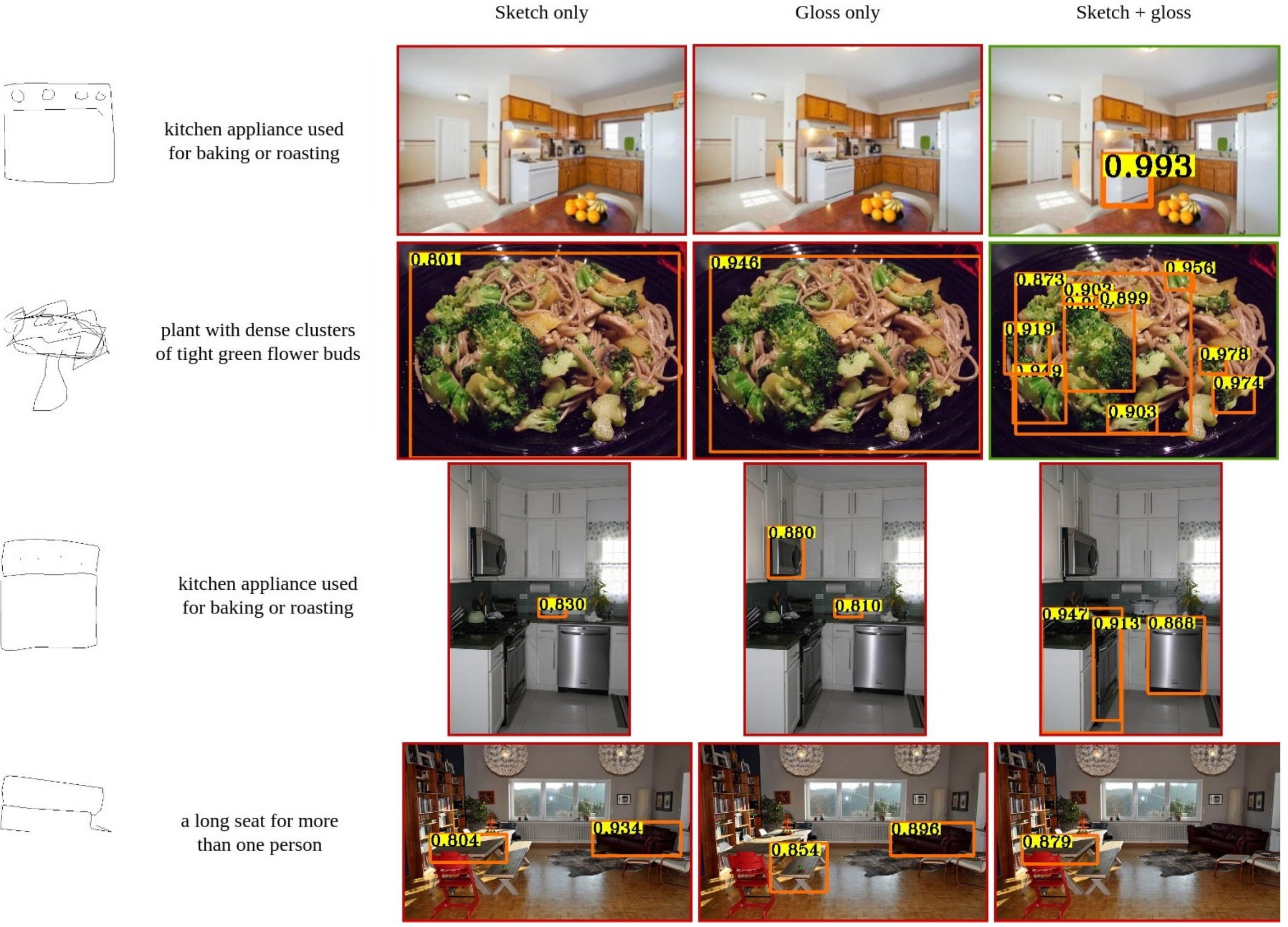}
    \caption{\label{fig:sketch_vs_gloss} The localization results are shown for the case when only sketch query (third column), only gloss query (fourth column), and both sketch and gloss queries (fifth column). The results are shown for the open-set setting, i.e., these categories are unseen during training. The first two columns show sketch and gloss queries. We observe that having gloss brings semantics to the model and thereby enables it to perform better than \emph{sketch only} localization. The last two rows show some of the failure cases.}
\end{figure*}

 \begin{figure*}[h!]
    \centering
    \includegraphics[width =\textwidth]{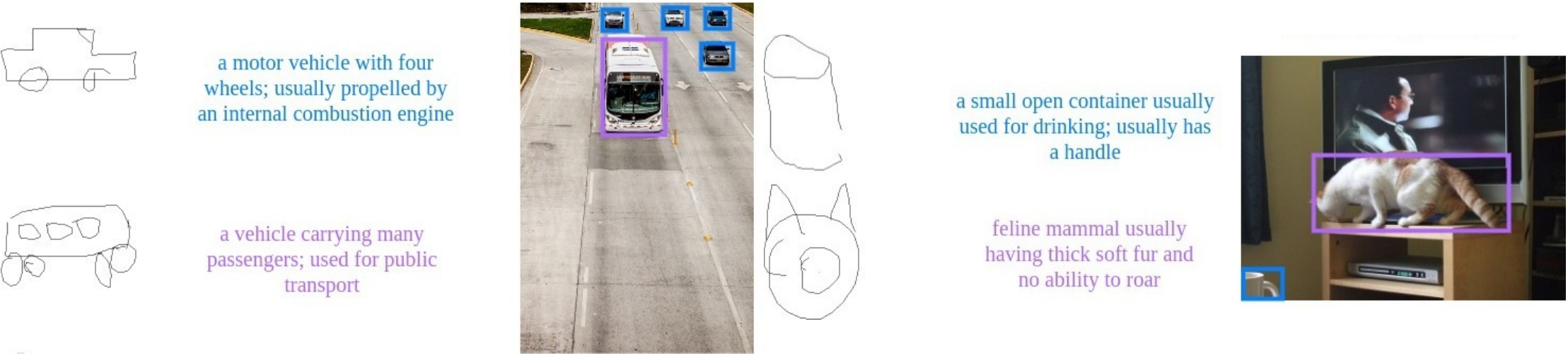}
    \caption{\label{fig:multitarget} The \textbf{multi-target localization results} are shown for the case when both sketch and gloss query are available. The results are shown for the open-set setting. The first column shows the sketch queries, and the second column shows the corresponding gloss queries using two different colors, i.e., blue and purple. Corresponding localizations in the target image are shown using the same colors as the gloss queries. \textbf{[Best viewed in color]}.}
\end{figure*}

\begin{table}[t]
\renewcommand{\arraystretch}{1.1}
\centering
\begin{tabular}{llll|lll}
\toprule
        & \multicolumn{3}{c|}{m} & \multicolumn{3}{c}{$\cal K$}   \\

        & 0.35    & 0.3     & 0.25   & 200   & 256   & 312     \\
\midrule
mAP     & 9.8     & 10.0    & 8.2    & 9.1   & 10.0  & 9.5     \\
\%AP@50 & 19.5    & 19.7    & 16.2   & 18.3  & 19.7  & 19.0    \\
\bottomrule

\end{tabular}
\caption{\label{tab:ablation_k} Effect of $m$ and $\cal K$ on the localization performance. The experiments are performed in \textit{MS-COCO} dataset.}
\end{table}
\subsubsection{\textbf{Ablation study}}
\textcolor{black}{Instead of the gloss of an object, the class name of an object can also be used as a query modality. We used class in two ways: i) append the object class at the beginning of the gloss and then use it as a query, and ii) use the word2vec embedding of the object class along with the object gloss and the sketch. In table~\ref{tab:ablation}, gloss with class refers to the case when the object category is appended in front of the gloss. The results in table~\ref{tab:ablation} suggest that using the class name, if available, helps in performance improvement. Moreover, using the word2vec embeddings of the class name along with the sketch and the gloss gives a $3.9\%$ improvement in performance. However, word2vec embeddings are trained on a large dataset set to learn semantic similarity between words, and therefore it violates the true open set experimental setting.}
 
\subsubsection{\textbf{Comparison across different dataset splits}} \textcolor{black}{In this experiment, we compared our proposed Orthogonal Proposal Scoring (OPS) with the Concatenation fusion on different splits of train and test categories. As shown in Table~\ref{tab:data_splits}, the performance of both of these methods varies across the splits, and our proposed OPS method performs the best across all the splits.}
 
\subsubsection{\textbf{Effect of $m$ and $\cal K$}} \textcolor{black}{In this experiment, we studied the effect of margin $m$ and scaling factor $\cal K$ on the localization performance of the model (Refer Table~\ref{tab:ablation_k}). The proposed model is fairly robust to changes in these parameters, and the best results are obtained when $m=0.3$ and {\cal K = 256}.}

\subsubsection{\textbf{Additional experiments}}
 Sketches are heterogeneous in quality, and they often tend to capture complementary information on an object's shape, characteristics, and appearance, and many times multiple sketches of an object can be utilized. Therefore, we also compare the proposed multimodal localization with multiple sketch-based localization, and in order to utilize multiple sketches, we use the following two fusion techniques~\citep{tripathi2020sketch}:
 
\noindent{\textbf{(i) Feature Fusion}}
\label{sec:query_fus}
Image feature maps for different sketch queries are first generated, and then global max pool operation is applied to fuse these feature maps. Let an image be queried by $N$ sketches that belong to the same object category $c\in C$, which is denoted as set $\{s_c^1,\dots, s_c^N\}$. These queries are then fed through the sketch backbone network, and suppose the representation learned for the $n^{th}$ sketch is denoted as ${^n}s_c^{\phi_S}$. These feature map representations for each query are concatenated together to yield a composite feature map $R^{w\times h\times d\times N}$. Finally, a global max pool operation is performed across all $N$ channels to obtain a fused feature map for the sketch queries.

\noindent{\textbf{(ii) Attention Fusion}}
\label{sec:atten_fus}
For each of the $N$ sketch queries, attention maps are first generated and concatenated. Then depth-wise mean pool operation is applied to obtain the resultant fused attention map, which is used as input to the object localization pipeline (Section \ref{sec:scoring}).

We perform an experiment where we use multiple (five) sketch queries instead of one, and these results are reported in Table~\ref{tab:multi_oneshot}. We observe that using multiple sketch queries improves the localization performance compared to using just a single sketch query. However, utilizing multiple modalities seems to perform better than fusing multiple sketch queries for both the open-set and closed-set experimental settings (refer to Tables~\ref{tab:few_shot_coco}). It could be attributed to the fact that the textual data are pretrained such that it captures the semantic similarity of the object categories and, therefore, captures complementary information from the visual representation.

\subsubsection{\textbf{Qualitative Results and Failure Analysis}}
To illustrate the effect of introducing the gloss as a query and the sketch, we visualize the localization results when the only sketch or gloss is available for the query and when both modalities are available. As shown in Fig.~\ref{fig:sketch_vs_gloss}, the gloss of an object is able to assist the sketch query to generate better localization for the case of open-set queries. These visualizations, along with the empirical results, illustrate that using semantic information from the gloss and shape information from the sketch helps improve the localization performance for unseen categories. \textcolor{black}{ In the fifth row, the model is not able to discriminate even when both the sketch and the gloss are available for query. Similarly, in the sixth row, the model is confused about the object represented by the queries and is only able to localize the part of the object. Moreover, when localization for either of the query is correct, the combined model is able to localize the object with better confidence.}

\textcolor{black}{Moreover, in Fig.~\ref{fig:multitarget}, we showed some qualitative results in which we query the same image with two different queries belonging to separate classes. As shown in the figure, the model is able to localize the correct objects.}


\subsubsection{\textbf{Comparison on Computational Time}}
\textcolor{black}{Our model is computationally efficient. On average, it takes 0.08 seconds per query (sketch + gloss) to localize objects in the target scene. Compared to this CoATex~\citep{hsieh2019one} and modified faster RCNN take 0.07 and 0.06 seconds per query, respectively, for localizing objects in the target scene. All these experiments are run on a system with Nvidia 1080-Ti GPU (with 11 GB RAM). 
}

\section{Conclusion}
\label{sec: conc}
In this paper, we have investigated multimodal query-guided object localization in natural images. Our proposed framework seamlessly fuses sketch and text queries and generates object proposals that are relevant to the query. We further proposed a novel proposal scoring mechanism using the orthogonal projection. 
The noticeable performance gain achieved over the baselines establishes the efficacy of the proposed framework. Moreover, the proposed framework, by virtue of query-guided proposal generation and our novel proposal scoring scheme, is also effective for open-set object localization. We have performed extensive experiments and demonstrated the utility of bringing semantics using gloss in the object localization framework. Our work further strengthens the argument in the literature, i.e., effectively using information across multiple modalities and exploiting their complementary nature can improve performance on learning tasks.

\section*{Acknowledgements}
The authors would like to thank the Advanced Data Management Research
Group, Corporate Technologies, Siemens Technology and Services Pvt. Ltd., Bengaluru, India, for partly supporting this research. Anand Mishra is supported by Startup Research Grant (SRG) by the SERB, Govt. of
India (File number: SRG/2021/001948).  

\bibliography{sn-bibliography}


\end{document}